\documentclass[final,5p,times,twocolumn]{elsarticle}

\usepackage{times}
\usepackage{amsfonts}
\usepackage{amsmath}
\usepackage{amssymb} 

\usepackage[nolist]{acronym}
\usepackage{xurl}

\usepackage{graphicx}
\usepackage{color}
\usepackage{bm}
\usepackage{siunitx}
\usepackage{multirow}
\usepackage{multicol}
\usepackage{makecell}

\usepackage{booktabs}
\usepackage{caption}

\usepackage{hyperref}
\hypersetup{
  colorlinks=true,
  allcolors=black
}

\usepackage{tikz}
\usepackage{tikz-3dplot}

\usepackage{tikz}
\definecolor{accessblue}{cmyk}{1, 0.3, 0, 0.2}
\definecolor{greycolor}{cmyk}{0,0,0,.8}
\usepackage{pgfplots}
\pgfplotsset{compat=1.14}
\usetikzlibrary{quotes, angles, backgrounds, arrows, automata, shapes, positioning, calc, through, spy, decorations.pathreplacing, decorations.markings, arrows.meta, automata, petri, intersections, pgfplots.fillbetween}
\pgfdeclarelayer{background}
\pgfdeclarelayer{foreground}
\pgfsetlayers{background, main, foreground}

\tikzset{
  state/.style={
    rectangle,
    draw=black, very thick,
    minimum height=1.0em,
    text centered,
  },
  smallstate/.style={
    rectangle,
    draw=black, very thick,
    minimum height=0.2em,
    text centered,
  },
  final_state/.style={
    rectangle,
    rounded corners,
    draw=black, very thick,
    minimum height=2em,
    text centered,
  },
  initial_state/.style={
    rectangle,
    double=white,
    double distance=1pt,
    inner sep=2pt,
    draw=black, very thick,
    minimum height=2em,
    text centered,
  },
  point/.style={
    circle,
    inner sep=0pt,
    minimum size=3pt,
    fill=red
  },
  adder/.style={
    circle,
    inner sep=2pt,
    minimum size=0.3in,
    draw=black, very thick,
    text centered
  },
  state_gray/.style={
    rectangle,
    draw=black, very thick,
    fill=gray!40,
    minimum height=1.0em,
    text centered,
    inner sep=0,
  },
  state_white/.style={
    rectangle,
    draw=black, very thick,
    fill=white,
    minimum height=1.0em,
    text centered,
    text=black,
    inner sep=0,
  },
  state_green/.style={
    rectangle,
    draw=black, very thick,
    fill=green!50,
    minimum height=1.0em,
    text centered,
    text=black,
    inner sep=0,
  },
  state_red/.style={
    rectangle,
    draw=black, very thick,
    fill=red!70,
    minimum height=1.0em,
    text centered,
    text=black,
    inner sep=0,
  },
  state_blue/.style={
    rectangle,
    draw=black, very thick,
    fill=blue!40,
    minimum height=1.0em,
    text centered,
    text=black,
    inner sep=0,
  },
  final_state/.style={
    rectangle,
    rounded corners,
    draw=black, very thick,
    minimum height=2em,
    text centered,
  },
  initial_state/.style={
    rectangle,
    double=white,
    double distance=1pt,
    inner sep=2pt,
    draw=black, very thick,
    minimum height=2em,
    text centered,
  },
  point/.style={
    circle,
    inner sep=0pt,
    minimum size=3pt,
    fill=red
  },
}

\tikzset{new spy style/.style={spy scope={
  magnification=5,
  size=1.25cm,
  connect spies,
  every spy on node/.style={
    rectangle,
    draw,
  },
  every spy in node/.style={
    draw,
    rectangle,
    fill=white
  }
  }
  }
  }

\newcommand{\reffig}[1]{Fig.~\ref{#1}}
\newcommand{\reffigfull}[1]{Figure~\ref{#1}}
\newcommand{\refsec}[1]{Sec.~\ref{#1}}
\newcommand{\reftab}[1]{Tab.~\ref{#1}}
\newcommand{\reftabfull}[1]{Table~\ref{#1}}
\newcommand{\refeq}[1]{\eqref{#1}}

\renewcommand{\vec}[1]{\mathbf{#1}}
\newcommand{\vioframe}[0]{\mathcal{V}}
\newcommand{\viobodyframe}[0]{\mathcal{B}_{v}}
\newcommand{\worldframe}[0]{\mathcal{W}}
\newcommand{\worldbodyframe}[0]{\mathcal{B}_{w}}
\newcommand{\dt}[0]{\text{d\textit{t}}}

\begin{document}

\begin{frontmatter}

\title{Onboard UAV State Estimation for Fast Flights Without External Localization Systems: A2RL Drone Racing Finalist Approach}
  \tnotetext[t1]{This work was funded by CTU grant no SGS23/177/OHK3/3T/13, by the Czech Science Foundation (GAČR) under research project no. 23-06162M, and by the European Union under the project Robotics and advanced industrial production (reg. no. CZ.02.01.01/00/22\_008/0004590).}

  \author{Filip Nov\'{a}k\corref{cor1}} 
  \ead{filip.novak@fel.cvut.cz}
  \author{Mat\v{e}j Petrl\'{i}k}
  \ead{matej.petrlik@fel.cvut.cz}
  \author{Matej Novosad}
  \ead{matej.novosad@fel.cvut.cz}
  \author{Parakh M. Gupta}
  \ead{guptapar@fel.cvut.cz}
  \author{Robert P\v{e}ni\v{c}ka}
  \ead{robert.penicka@fel.cvut.cz}
  \author{Martin Saska}
  \ead{martin.saska@fel.cvut.cz}

  \cortext[cor1]{Corresponding author.}

  \affiliation{organization={Department of Cybernetics, Faculty of Electrical Engineering, Czech Technical University in Prague},
            postcode={166 36}, 
            city={Prague 6},
            country={Czech Republic},
            }

\begin{abstract}
Fast flights with aggressive maneuvers in cluttered GNSS-denied environments require fast, reliable, and accurate UAV state estimation.
In this paper, we present an approach for onboard state estimation of a high-speed UAV using a monocular RGB camera and an IMU.
Our approach fuses data from Visual–Inertial Odometry (VIO), an onboard landmark-based camera measurement system, and an IMU to produce an accurate state estimate.
Using onboard measurement data, we estimate and compensate for VIO drift through a novel mathematical drift model.
State-of-the-art approaches often rely on more complex hardware (e.g., stereo cameras or rangefinders) and use uncorrected drifting VIO velocities, orientation, and angular rates, leading to errors during fast maneuvers.
In contrast, our method corrects all VIO states (position, orientation, linear and angular velocity), resulting in accurate state estimation even during rapid and dynamic motion.
Our approach was thoroughly validated through 1600 simulations and numerous real-world experiments.
Furthermore, we applied the proposed method in the A2RL Drone Racing Challenge 2025, where our team advanced to the final four out of 210 teams and earned a medal.
\end{abstract}


\end{frontmatter}

\section*{SUPPLEMENTARY MATERIALS}
\noindent{\small\textbf{Video:} \href{https://youtu.be/7729hzU34aE}{https://youtu.be/7729hzU34aE}}\\
\noindent{\small\textbf{Web:} \href{https://mrs.fel.cvut.cz/papers/onboard-uav-state-estimation-a2rl}{https://mrs.fel.cvut.cz/papers/onboard-uav-state-estimation-a2rl}}\\
\noindent{\small\textbf{Code:} \href{https://github.com/ctu-mrs/onboard-uav-state-estimation-a2rl}{https://github.com/ctu-mrs/onboard-uav-state-estimation-a2rl}}

\section{INTRODUCTION}

Agile lightweight drones have gained significant attention in recent years, due to their potential ability to fly fast even in a highly cluttered environment.
These capabilities make them well-suited for inspection tasks \cite{power_line_inspection}, search-and-rescue operations \cite{petrlik2023uavs}, or security and surveillance missions \cite{water_surface_monitoring}.
Autonomous \acp{UAV} typically operate at low speeds near hover, ensuring adequate time for their sensing systems to collect data to estimate the robot's state within the environment, which is not efficient for the applications mentioned above.
Pushing autonomous \acp{UAV} towards aggressive maneuvers at high speeds with acceleration up to \SI{7}{g} in cluttered \ac{GNSS}-denied environments, which is tackled in this paper, places significant demands on \ac{UAV} state estimation, especially when \ac{UAV} payload capacity for onboard sensors is strictly limited and no external localization system, e.g., a~motion capture system, is available.
Robust state estimation under such extreme conditions has become a main topic of interest for many research studies \cite{delmerico2019AreWeReady, 
kaufmann2019BeautyBeastOptimal,
foehn2022AlphaPilotAutonomousDrone,
kaufmann2023ChampionlevelDroneRacing, azhari2025driftcorrectedmonocularvioperceptionaware, bahnam2026monoracewinningchampionleveldrone, azhari2026robusttightlycoupledfilterbasedmonocular, xu2021CNNbasedEgoMotionEstimation}.

\begin{figure}[!t]
  \centering
\includegraphics[width=\linewidth]{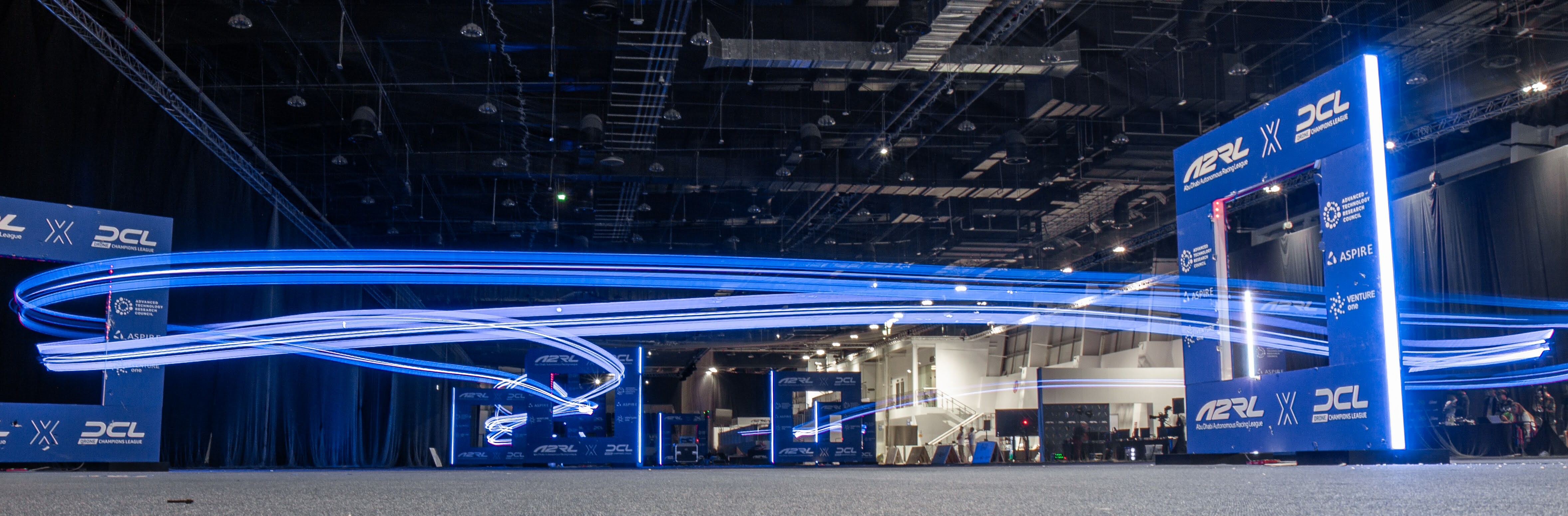}
  \caption{Long-exposure photo of drones carrying blue lights taken during a race flight at the A2RL Drone Racing Challenge 2025.}
  \label{fig:flight_long_exposure}
\end{figure}

To stress-test and to improve autonomous agility and state estimation under these challenging environmental and sensory constraints, high-speed competitions such as the A2RL Drone Racing Challenge\footnote{\href{https://a2rl.io/autonomous-drone-race}{https://a2rl.io/autonomous-drone-race}\label{a2rl_challenge}} have emerged as rigorous benchmarking avenues (see \reffig{fig:flight_long_exposure}).
The competition restricted each \ac{UAV} to a monocular RGB camera, an \ac{IMU}, and a lightweight onboard computer while requiring aggressive flight through a predefined sequence of gates at maximum speed.
While typical competition solutions rely on end-to-end learning or reinforcement learning methods \cite{kaufmann2023ChampionlevelDroneRacing, bahnam2026monoracewinningchampionleveldrone, azhari2026robusttightlycoupledfilterbasedmonocular} that are often overfitted to the specific racing track, these solutions lack the generalizability required for broader deployment in industrial or safety-critical applications, such as infrastructure inspection \cite{power_line_inspection} and search-and-rescue operations \cite{petrlik2023uavs}.
In contrast, our approach treats this high-speed setup as a validation testbed for generalized flight.
We decompose the problem into modular planning, state estimation, and a \ac{MPC} framework, where Lyapunov stability theory provides formal guarantees \cite{Zheng2024StabilityGuaranteedMPC} that learning-based alternatives typically lack.

The state-of-the-art methods rely on more complex hardware, such as stereo cameras and rangefinders \cite{foehn2022AlphaPilotAutonomousDrone}, or they depend on \ac{UAV} states provided by \ac{VIO} \cite{foehn2022AlphaPilotAutonomousDrone, kaufmann2023ChampionlevelDroneRacing, azhari2025driftcorrectedmonocularvioperceptionaware}, which fuses data from a single RGB camera and an \ac{IMU}.
However, \ac{VIO} suffers from drifts, inaccuracies, and slow reaction to aggressive changes in all \ac{UAV} states. 
The method presented in \cite{foehn2022AlphaPilotAutonomousDrone} uses uncorrected linear and angular velocities, which cause significant issues for the \ac{UAV} controller.
Similarly, the approach \cite{kaufmann2023ChampionlevelDroneRacing} relies on uncorrected attitude and angular rates, leading to problems for controller and onboard vision-based systems.

Our approach running at \SI{100}{\hertz} leverages \ac{VIO} \cite{qin2018VINSMonoRobustVersatile}, but compared to the state-of-the-art methods \cite{foehn2022AlphaPilotAutonomousDrone, kaufmann2023ChampionlevelDroneRacing}, we correct all \ac{UAV} states, including position, orientation, linear velocity and angular velocity.
We also use our onboard landmark-based camera measurement system, which uses known landmarks in an environment that are detectable in camera images, to provide an estimate of the \ac{UAV}'s position and orientation.
However, any other method that provides such noisy partial \ac{UAV} state estimates within the operating area can also be used.

In order to reduce \ac{VIO} translational drift as well as attitude and velocity inaccuracies, our method fuses the estimated states obtained from detections of known landmarks with \ac{VIO} and \ac{IMU} data.
We introduce a novel model of \ac{VIO} drift that includes translation, linear velocity, yaw angle, and yaw rate, where the velocities are influenced by artificial friction to increase accuracy even when the onboard camera system temporarily stops providing usable data.
Additionally, \ac{IMU} data are directly fused into the attitude estimation in our pipeline, enabling higher accuracy and faster response to rapid maneuvers compared to state-of-the-art methods \cite{foehn2022AlphaPilotAutonomousDrone, kaufmann2023ChampionlevelDroneRacing}.
We reduced orientation \ac{RMSE} by \SI{57}{\percent}, linear velocity \ac{RMSE} by \SI{18}{\percent}, and angular velocity \ac{RMSE} by factor of 9 compared to \cite{kaufmann2023ChampionlevelDroneRacing} and \cite{foehn2022AlphaPilotAutonomousDrone}.
The presented estimation framework was also successfully deployed at the A2RL Drone Racing Challenge 2025\textsuperscript{\ref{a2rl_challenge}}, where we reached the finals among the top four teams and won a medal, achieving top performance among classical approaches with a model-based state estimation and control pipeline suitable for industrial and safety-critical applications.

\begin{figure*}
  \centering
  \resizebox{1.0\textwidth}{!}{
   \input{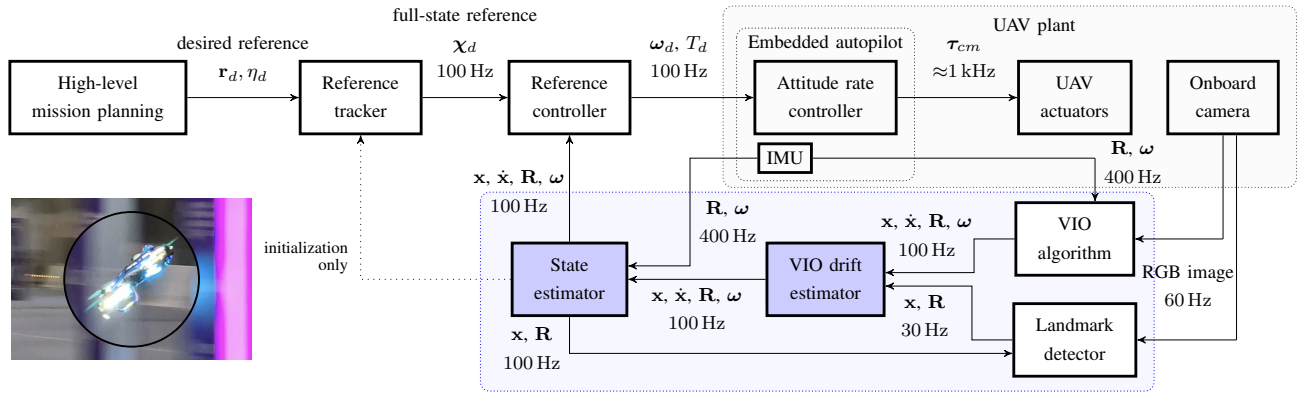}}
    \caption{
    This figure presents the pipeline diagram of the estimation approach proposed in this paper, highlighted by the light blue dotted rectangle and integrated into our drone racing framework. 
    The \textit{High-level mission planning} block provides the desired position and heading references $(\vec{r}_d,\eta_d)$ for the UAV to follow the race track. 
    These references are passed to the \textit{Reference tracker}, which generates a smooth and dynamically feasible reference $\bm{\chi}_d$ for the \textit{Reference controller}.
    The \textit{Reference controller} computes the desired thrust and angular velocities $(\bm{\omega}_d,~T_d)$ for the embedded flight controller, which then sends actuator commands $\bm{\tau}_{cm}$ to the \textit{UAV actuators}.
    The \textit{VIO algorithm} uses \textit{IMU} data and onboard camera images to estimate the UAV position $\vec{x}$, orientation $\vec{R}$, linear velocity $\dot{\vec{x}}$, and angular velocity $\bm{\omega}$.
    The \textit{VIO drift estimator} estimates the drift in the \ac{VIO} states using position $\vec{x}$ and orientation $\vec{R}$ measurements from the \textit{Landmark detector}.
    The \textit{State estimator} fuses data from the \textit{IMU}, \textit{VIO drift estimator}, and \textit{VIO algorithm} to estimate the UAV position, orientation, and linear and angular velocities $(\vec{x},\dot{\vec{x}},\mathbf{R},\bm{\omega})$.
    }
    \label{fig:diagram}
\end{figure*}

\section{RELATED WORKS}
Recent research has explored \ac{UAV} state estimation using known landmarks and onboard \acp{IMU} \cite{bahnam2026monoracewinningchampionleveldrone, azhari2026robusttightlycoupledfilterbasedmonocular}.
These methods detect racing gates in camera images and associate the detections with a known map of gates to estimate the \ac{UAV} position.
The position estimate is then fused with \ac{IMU} measurements to recover the complete \ac{UAV} state.
However, the performance of these methods degrades rapidly when gates are not visible within the camera's \ac{FOV}.
Analogously, in practical scenarios such as inspection and surveillance, the desired landmarks are not always visible in the camera images.
Furthermore, these approaches are highly task-specific and do not generalize well to environments or applications beyond those for which they were originally designed.
Several studies have adopted deep learning techniques for indoor state estimation using a monocular camera combined with an \ac{IMU} \cite{wang2021TartanVOGeneralizableLearningbased, xu2021CNNbasedEgoMotionEstimation, xu2025CUAHNVIOContentanduncertaintyawareHomographya}.
These methods provide end-to-end 6 \acp{DOF} \ac{UAV} state estimation from image sequences fused with an \ac{IMU}.
However, these methods require large training datasets, are computationally intensive, and generalize poorly to unseen environments.

\ac{VIO} represents the typical method for state estimation in \ac{GNSS}-denied environments, running onboard autonomous robots equipped with a monocular camera and \ac{IMU} \cite{scaramuzza2020AerialRobotsVisualInertial, qin2018VINSMonoRobustVersatile}.
\ac{VIO} estimates camera motion from image sequences, a process known as \ac{VO} method, which is categorized into direct and feature-based methods.
Direct methods \cite{engel2018DirectSparseOdometry, usenko2016DirectVisualinertialOdometry} optimize photometric error defined over raw pixel-level intensities.
They handle low-texture environments but are sensitive to initialization and illumination changes.
In contrast, feature-based \ac{VO} methods \cite{geneva2020OpenVINSResearchPlatform, qin2018VINSMonoRobustVersatile} detect visual features in images and track them across successive image frames.
Feature-based methods are generally more robust to initialization and illumination changes but tend to be more affected by motion blur compared to direct methods.

The performance and accuracy of the \ac{VO} approaches are enhanced by fusing \ac{IMU} measurements, leading to \ac{VIO} methods.
\acp{IMU} typically provide data at a much higher frequency than cameras. 
Therefore, the integration of \ac{IMU} significantly improves estimation of \ac{UAV}'s attitude and motion tracking in cases of losing visual features \cite{usenko2016DirectVisualinertialOdometry, geneva2020OpenVINSResearchPlatform, qin2018VINSMonoRobustVersatile}.
Despite these improvements, all aforementioned \ac{VIO} methods still suffer from drift, where the estimated position, orientation, and linear and angular velocity gradually deviate from ground-truth values over time, especially during fast flights with high motion blur \cite{delmerico2019AreWeReady}.
Furthermore, \ac{UAV} attitude estimates may lack precision and often respond slowly to rapid changes in orientation.

Researchers have also investigated the use of event cameras in \ac{VIO} systems \cite{vidal2018UltimateSLAMCombining, chen2023ESVIOEventBasedStereo, guan2024PLEVIORobustMonocular}.
Event cameras aim to address key limitations of standard cameras, such as fixed frame rates, low dynamic range, and motion blur, which pose significant challenges for \ac{VIO} algorithms.
Unlike standard cameras, event cameras offer low latency, high dynamic range, and are inherently immune to motion blur.
However, event-based \ac{VIO} methods are still in the early stages of development and continue to suffer from drift.
Other approaches incorporate vehicle dynamics and external force estimation into \ac{VIO} frameworks \cite{nisar2019VIMOSimultaneousVisual, ding2021VIDFusionRobustVisualInertialDynamics, cioffi2023HDVIOImprovingLocalization}, improving state estimation performance, especially at high speeds, but issues related to drift and inaccurate velocity estimation persist.

The approach presented in \cite{foehn2022AlphaPilotAutonomousDrone} employs a \ac{VIO} method fused with measurements from a vision-based gate detector and a downward-facing laser rangefinder to achieve accurate state estimation in \ac{GNSS}-denied environments.
Gate detections serve as features used to compensate \ac{VIO} drift and align the \ac{UAV} with its intended path.
A Kalman filter estimates the translational and rotational misalignment between the \ac{VIO} frame and the inertial frame.
However, this approach cannot compensate for inaccuracies in the estimated \ac{VIO} velocities, which may cause significant issues for controllers, particularly during high-speed aggressive maneuvers.
Furthermore, the use of a laser rangefinder and stereo cameras for \ac{VIO} simplifies the problem compared to our scenario, which relies on a single monocular camera.

The approach \cite{kaufmann2023ChampionlevelDroneRacing} fuses only \ac{VIO} estimates with gate detections.
The method uses a known race track layout to estimate the \ac{UAV} position based on detected gates.
The position estimates derived from gate measurements are used to correct \ac{VIO} translational and linear velocity drift, which is modeled as a simple double integrator with zero input.
However, this method strongly relies on attitude and angular rate estimates provided by \ac{VIO}.
During high-speed aggressive maneuvers, \ac{VIO} estimates of attitude and angular rates are also subject to drift and latency. 
Such an inaccurate attitude estimates introduce significant errors in onboard camera-based measurement systems.
Method \cite{azhari2025driftcorrectedmonocularvioperceptionaware} follows a similar approach to \cite{kaufmann2023ChampionlevelDroneRacing}. However, only the VIO position is corrected, whereas all other states are obtained directly from the \ac{VIO} estimates and therefore remain affected by drift and estimation errors, especially during fast and aggressive flights.

In contrast to the most relevant state-of-the-art methods \cite{foehn2022AlphaPilotAutonomousDrone, kaufmann2023ChampionlevelDroneRacing}, our approach: (1) compensates for translational, rotational, and both linear and angular velocity drifts in the \ac{VIO} state estimates,
(2) introduces a novel drift model, (3) directly fuses \ac{IMU} data into the final \ac{UAV} state estimate, enabling faster and more accurate responses to aggressive attitude changes, and (4) addresses the crucial issue of outliers arising in data from the onboard vision-based measurement system.
Furthermore, our extensive experimental analysis (\refsec{sec:results}) demonstrates that our approach outperforms state-of-the-art methods \cite{foehn2022AlphaPilotAutonomousDrone, kaufmann2023ChampionlevelDroneRacing}, reducing orientation \ac{RMSE} by \SI{57}{\percent}, linear velocity \ac{RMSE} by \SI{18}{\percent}, and angular velocity \ac{RMSE} by a factor of 9.
Our method is compatible with any direct or feature-based \ac{VIO} method presented above that provides position, orientation, yaw, and yaw rate. 
If our approach is deployed within a feedback control loop, the underlying \ac{VIO} method is required to operate at a minimum frequency of \SI{100}{\hertz}.
Additionally, our method can integrate with arbitrary vision-based measurement systems that provide the \ac{UAV} position, velocity, yaw angle, and yaw rate.
The source code is released as open-source\footnote{\href{https://github.com/ctu-mrs/onboard-uav-state-estimation-a2rl}{https://github.com/ctu-mrs/onboard-uav-state-estimation-a2rl}}.

\section{METHODOLOGY}
The pipeline diagram of the proposed system is illustrated in \reffig{fig:diagram}.
Our approach uses the \ac{VIO} algorithm \cite{qin2018VINSMonoRobustVersatile}, which utilizes onboard \ac{IMU} and RGB camera data.
The \ac{VIO} module provides drifting and inaccurate \ac{UAV} state estimates to the \ac{VIO} drift estimator, where the drift is compensated using additional data from our onboard landmark-based camera measurement system, i.e., Landmark detector, that provides estimated position and orientation of the \ac{UAV} at \SI{30}{Hz} based on detection of gates with known position.
The corrected states from the \ac{VIO} drift estimator are subsequently fused with the onboard high-rate \ac{IMU} measurements (\SI{400}{\hertz}) in the State estimator block (see \reffig{fig:diagram}), yielding full-state estimates that are provided to the Reference controller \cite{gupta2024lolnmpc} at \SI{100}{\hertz}.
The Reference controller tracks the desired full-state reference produced by the Reference tracker, which follows a trajectory precomputed using the minimum-time trajectory generation method for the point-mass model \cite{teissing2024pmm}.
Finally, the Reference controller transmits the control commands to the embedded autopilot, which actuates the \ac{UAV}'s motors accordingly.

\begin{figure}[!bt]
      \centering
      \includegraphics[width=\linewidth]{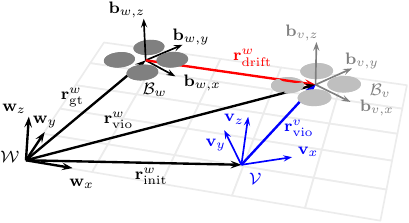}
      \caption{
      The depiction of coordinate frames consisting of
      the world frame $\worldframe{}$, \ac{UAV} body frame $\worldbodyframe{}$ expressed in the $\worldframe{}$ frame, \ac{VIO} frame $\vioframe{}$, and \ac{UAV} body frame $\viobodyframe{}$ expressed in the $\vioframe{}$ frame.
      The relative position and orientation between the $\worldframe{}$ frame and the $\vioframe{}$ frame are represented by $\vec{r}_{\text{init}}^w$.
      The \ac{UAV} position and orientation in $\worldframe{}$ frame are denoted as $\vec{r}_{\text{gt}}^w$.
      The \ac{UAV} position and orientation estimated by the \ac{VIO} algorithm are provided in the $\vioframe{}$ frame and denoted as $\vec{r}_{\text{vio}}^v$.
      The vector $\vec{r}_{\text{vio}}^w$ represents the \ac{VIO}-estimated \ac{UAV} position and orientation expressed in the $\worldframe{}$ frame.
      The \ac{VIO} drift is expressed as $\vec{r}_{\text{drift}}^w$.
      }
      \label{fig:coordinate_frames}
\end{figure}

In our approach, we estimate the full 6 \ac{DOF} \ac{UAV} state
\begin{align}
    \vec{x}_{\text{uav}} = (x,y,z,\phi,\theta,\psi,\dot{x}, \dot{y}, \dot{z},p,q,r)^\intercal,
\end{align}
where $\mathbf{p}_{\text{uav}}=(x,y,z)^\intercal$ denotes the position in the world frame $\worldframe{}$, $\mathbf{\Theta}_{\text{uav}}=(\phi,\theta,\psi)^\intercal$ represents the orientation in the $\worldframe{}$ frame in terms of roll, pitch, and yaw angles, $\mathbf{v}_{\text{uav}}=(\dot{x},\dot{y},\dot{z})^\intercal$ is the linear velocity in the $\worldframe{}$ frame, and $\bm{\omega}_{\text{uav}}=(p,q,r)^\intercal$ corresponds to the angular velocity in the \ac{UAV} body frame $\worldbodyframe{} = \{ \mathbf{b}_{w,x}, \mathbf{b}_{w,y}, \mathbf{b}_{w,z} \}$.
The \ac{UAV}'s position and orientation from \ac{VIO} algorithm $\mathbf{r}_{\text{vio}}^{v}=(\mathbf{p}_{\text{vio}}^\intercal, \mathbf{\Theta}_{\text{vio}}^\intercal)^\intercal$ are provided in a local reference frame $\vioframe{}$ defined at the initialization point of the \ac{VIO} method, as illustrated in \reffig{fig:coordinate_frames}.
The \ac{VIO} linear velocity $\mathbf{v}_{\text{vio}}$ and angular velocity $\bm{\omega}_{\text{vio}}$ of the \ac{UAV} are expressed in the \ac{UAV} body frame $\viobodyframe{} = \{ \mathbf{b}_{v,x}, \mathbf{b}_{v,y}, \mathbf{b}_{v,z} \}$.
The \ac{UAV} states in the $\vioframe{}$ frame are transformed into the $\worldframe{}$ using a static transformation $\boldsymbol{T}_{\vioframe{}}^{\worldframe{}}$, which is constructed from the known initial \ac{UAV} position and orientation $\mathbf{r}_{\text{init}}^{w}$ in the $\worldframe{}$ frame
\begin{align}
    \mathbf{r}_{\text{vio}}^{w} = \boldsymbol{T}_{\vioframe{}}^{\worldframe{}} (\mathbf{r}_{\text{vio}}^{v}).\label{eq:vio_w_frame}
\end{align}
The ground-truth position and orientation of the \ac{UAV} in the $\worldframe{}$ frame are denoted as $\mathbf{r}_{\text{gt}}^{w}$.
The ground-truth linear and angular velocities are defined in the \ac{UAV} body frame $\worldbodyframe{} = \{ \mathbf{b}_{w,x}, \mathbf{b}_{w,y}, \mathbf{b}_{w,z} \}$.
The position and orientation \ac{VIO} drift $\mathbf{r}_{\text{drift}}^{w}$ in the $\worldframe{}$ frame is defined by the difference between the VIO states $\mathbf{r}_{\text{vio}}^{w}$ and the ground-truth states $\mathbf{r}_{\text{gt}}^{w}$.
The position $\vec{p}_{\text{ld}}^{w}$ and orientation $\vec{\Theta}_{\text{ld}}^{w}$ measured by the onboard landmark-based camera system are provided in the $\worldframe{}$ frame.

\subsection{VIO algorithm}
\label{sec:vio_algorithm}
The VINS-Mono \cite{qin2018VINSMonoRobustVersatile} \ac{VIO} algorithm is used as the primary inaccurate and drifting source of \ac{UAV} states.
However, the proposed method is agnostic to the choice of \ac{VIO} algorithm and was also successfully deployed with the OpenVINS \ac{VIO} algorithm \cite{geneva2020OpenVINSResearchPlatform}.
These \ac{VIO} methods require only a single camera and an \ac{IMU} to estimate the full 6 \ac{DOF} robot state.
To ensure smooth state feedback and conserve computational resources, we intentionally disabled the loop closure and pose graph optimization modules. This avoids sudden jumps in state estimates that could degrade control performance, prevents interference with our drift compensation, and reduces onboard processing overhead.
The VINS-Mono algorithm \cite{qin2018VINSMonoRobustVersatile} outputs the estimated states in the $\vioframe{}$ frame, which is established during initialization.
The initialization procedure estimates key parameters, including metric scale, gravity direction, and gyroscope bias.

Although VINS-Mono estimates the gravity vector during initialization, inaccuracies in this process lead to noticeable errors in roll and pitch.
These errors occur because the gravity direction is inferred from short-term motion and \ac{IMU} preintegration, both of which are sensitive to sensor noise, bias instability, and insufficient motion excitation during initialization. 
Consequently, the estimated gravity vector may deviate from the true vertical, resulting in a tilt of the $\vioframe{}$ frame.
To address this issue, we positioned the drone on a level starting podium with known position after \ac{VIO} initialization, and then performed a calibration procedure.
This procedure corrects the initial source of drift in \ac{VIO} odometry by resetting the \ac{UAV}'s position, roll, and pitch to zero, thereby effectively re-aligning the $\vioframe{}$ frame with the true gravity direction
\begin{align}
    \vec{p}_{\text{vio}}(t) &= \vec{p}_{\text{vio},u}(t) - \vec{p}_{\text{vio},u}(t_{\text{init}}),\\
    \mathbf{q}_{\text{vio}}(t) &= \mathbf{q}_{\text{vio},u}^{-1}(t_{\text{init}}) \otimes \mathbf{q}_{\text{vio},u}(t),
\end{align}
where $\vec{p}_{\text{vio}}(t)$ is the calibrated \ac{VIO} position at time $t$, $\vec{p}_{\text{vio},u}(t)$ is the uncalibrated \ac{VIO} position at time $t$, $\vec{p}_{\text{vio},u}(t_{\text{init}})$ is the uncalibrated \ac{VIO} position at time $t_{\text{init}}$ when the calibration procedure is invoked, $\mathbf{q}_{\text{vio}}(t)$ is the calibrated \ac{VIO} orientation of the \ac{UAV} expressed as a quaternion, $\mathbf{q}_{\text{vio},u}(t_{\text{init}})$ is the uncalibrated orientation quaternion at time $t_{\text{init}}$, $\mathbf{q}_{\text{vio},u}(t)$ is the uncalibrated orientation quaternion provided by the \ac{VIO} algorithm, and $\otimes$ denotes the quaternion multiplication.
By ensuring that the \ac{UAV} starts from a known, level orientation, this procedure effectively removes accumulated attitude and position offsets, leading to more accurate and stable state estimation during operation.
To handle scenarios without a level starting podium — or when the podium's position and heading are inaccurate — our system utilizes landmarks during the preflight phase.
While stationary prior to takeoff, the filter employs a static model where velocities are omitted, allowing the landmark detector to directly correct the initial position and orientation states.
Upon takeoff, velocity estimation is activated.
This enables the system to initialize reliably on arbitrary or inclined surfaces without requiring a pre-mapped launch platform.

\subsection{Onboard landmark-based camera measurement system}
\label{sec:gate_estimator}
To correct the drifting odometry from the \ac{VIO} pipeline, we utilize an onboard landmark-based camera measurement system, referred to as Landmark detector (\reffig{fig:diagram}).
The Landmark detector provides estimated \ac{UAV} states in the $\worldframe{}$ frame using predefined static landmarks.
In our scenario, we use racing gates as static landmarks with known positions and orientations.
We also utilize the known pose of the starting podium as an initial state estimate, enabling the association of detected gate candidates with their real-world counterparts.
The Landmark detector computes the \ac{UAV}'s position $\vec{p}_{\text{ld}}^{w}$ and orientation $\vec{\Theta}_{\text{ld}}^{w}$ in the $\worldframe{}$ frame using a monocular RGB camera and the most recent estimated \ac{UAV} states.
The system operates in real time by combining deep learning–based corner detection, geometric reasoning, and nonlinear optimization to estimate the \ac{UAV} states relative to known gate locations in the environment.
Since the Landmark detector is not the primary focus or contribution of this paper, we provide only a brief description, as our proposed method uses its outputs for sensor fusion.

The Landmark detector processes RGB images from the \ac{UAV}’s onboard camera using a convolutional neural network based on the U-Net architecture \cite{foehn2022AlphaPilotAutonomousDrone}.
The network adopts an encoder–decoder structure with skip connections, consisting of five convolutional blocks (32, 64, 128, 256, and 512 channels) for feature extraction and transposed convolutions for upsampling.
Each block utilizes $3\times3$ kernels with padding to preserve spatial dimensions.
The final $1\times1$ convolution layer outputs 12 feature maps: 4 heatmaps corresponding to corner keypoints and 8 maps representing two-dimensional vector fields that describe the 4 gate edges.

Detected corners and edges are first used to assemble gate candidates.
These detected gate candidates are then matched to known real-world gate positions based on the predefined track layout and the \ac{UAV}'s most recent estimated pose.
Subsequently, the Landmark detector optimizes the \ac{UAV} states $\vec{p}_{\text{ld}}^w$ and $\vec{\Theta}_{\text{ld}}^w$ by minimizing the difference between optical rays projected from detected gate corners in the camera images and the corresponding real-world 3D positions of the gate corners.
The optimization is initialized using the most recent estimated \ac{UAV} position and orientation $\vec{r}_{\text{uav}}^w$.

Owing to imperfect calibration and variable camera-\ac{IMU} delays (see \reftab{tab:rmse_sim}), roll $\phi_{\text{ld}}$ and pitch $\theta_{\text{ld}}$ estimates from the Landmark detector are highly unstable and noisy. Because roll and pitch exhibit fast rotational dynamics, even small time delays induce substantial orientation errors. Conversely, because yaw dynamics $\psi_{\text{ld}}$ are slower due to motor torque limits around the $z$-axis, yaw estimation is slightly more robust to delays. Therefore, we fuse position and heading from the Landmark detector, defined as $\vec{x}_{\text{ld}}^w = (x_{\text{ld}}, y_{\text{ld}}, z_{\text{ld}}, \psi_{\text{ld}})^\intercal$.

The reprojection error $e_r$ measures the discrepancy between the optical rays from detected gate corners and their corresponding real-world 3D positions. 
This error serves as the cost metric and is converted into a confidence score $c_{\text{ld}}$ using a Gaussian weighting function
\begin{align}
    c_{\text{ld}} = \exp\Bigg(- \frac{e_{r}^2}{2 \sigma^2} \Bigg) \Big(1 - \exp(-\beta N_p)\Big),\label{eq:confidence}
\end{align}
where $e_{r}$ is the reprojection error in pixels, $\sigma$ is set to 0.15, $\beta$ is 0.4, and $N_p$ is the number of gate corners used in optimization to determine the \ac{UAV} state.
The hyperparameters $\sigma$ and $\beta$ were determined experimentally through a grid-search sweep on a separate validation dataset, optimizing for minimum state estimation drift.
The estimated \ac{UAV} state $\vec{x}_{\text{ld}}^w$ and its associated confidence $c_{\text{ld}}$ are then fused by our estimation pipeline as shown in \reffig{fig:diagram}.

\subsection{VIO drift estimation}
We model the drift as a discrete linear system 
\begin{align}
    \vec{x}_d(t+\dt{}) = \vec{A}_d\vec{x}_d(t) + \vec{w}_{\text{noise}}(t), 
\end{align}
where $t$ is the current time, the drift state $\vec{x}_d$ consists of the translational drift $\vec{p}_d = (x_d, y_d, z_d)^\intercal$ with corresponding drift velocities $\vec{v}_d = (\dot{x}_d, \dot{y}_d, \dot{z}_d)^\intercal$, as well as the yaw drift $\psi_d$ with its corresponding drift rate $r_d$
\begin{align}
    \vec{x}_d &= \begin{pmatrix}
        x_d & \dot{x}_d & y_d & \dot{y}_d & z_d & \dot{z}_d & \psi_d & r_d
    \end{pmatrix}^\intercal,\\[0.2cm] 
    \vec{A}_d &=
    \begin{pmatrix}
        1 & \dt{} & 0 & 0 & 0 & 0 & 0 & 0\\
        0 & 1-f_x & 0 & 0 & 0 & 0 & 0 & 0\\
        0 & 0 & 1 & \dt{} & 0 & 0 & 0 & 0\\
        0 & 0 & 0 & 1-f_y & 0 & 0 & 0 & 0\\
        0 & 0 & 0 & 0 & 1 & \dt{} & 0 & 0\\
        0 & 0 & 0 & 0 & 0 & 1-f_z & 0 & 0\\
        0 & 0 & 0 & 0 & 0 & 0 & 1 & \dt{}\\
        0 & 0 & 0 & 0 & 0 & 0 & 0 & 1-f_\psi
    \end{pmatrix}.
\end{align}
The parameter \dt{} represents the discrete time update step.
Each drift velocity component ($\dot{x}_d$, $\dot{y}_d$, $\dot{z}_d$) and yaw drift rate $r_d$ is influenced by the corresponding friction parameter $f_x$, $f_y$, $f_z$, $f_\psi$, which limit the rate of drift change.
Since the estimated drift velocity is only accurate over a short time interval, the system relies on this estimate for a limited period after visual landmarks are lost.
This period is governed by the friction parameters and depends on the properties of the hardware platform and the employed \ac{VIO} algorithm.
In practice, the friction parameters are tuned based on the maximum expected duration without landmark detections.
Since the \ac{VIO} drift cannot be directly affected or corrected through known control inputs, the model does not include any control input term.
However, in our drift model, we incorporate an unknown input vector $\vec{w}_{\text{noise}} \in \mathbb{R}^8$ to account for unmodeled dynamics, disturbances, and uncertainties.
Each component of $\vec{w}_{\text{noise}}$ is modeled as independent zero-mean Gaussian noise with covariance values $Q_{x_d}$, $Q_{y_d}$, $Q_{z_d}$, $Q_{\psi_d}$, $Q_{\dot{x}_d}$, $Q_{\dot{y}_d}$, $Q_{\dot{z}_d}$, $Q_{r_d}$ $\in \mathbb{R}^+$ corresponding to the individual drift states.

The \ac{VIO} drift cannot be directly measured or observed.
To estimate \ac{VIO} drift, we utilize the Landmark detector (\refsec{sec:gate_estimator}) that provides a noisy estimate of the \ac{UAV} state $\vec{x}_{\text{ld}}^w$ in the $\worldframe{}$ frame.
The position and yaw drift measurement $\vec{z}$ is then computed as the difference between the \ac{VIO} position and yaw $(x_{\text{vio}},y_{\text{vio}},z_{\text{vio}},\psi_{\text{vio}})$ expressed in $\worldframe{}$ frame \refeq{eq:vio_w_frame} and the camera-based estimate $\vec{x}_{\text{ld}}^w$
\begin{align}
    \vec{z} = (x_{\text{vio}},y_{\text{vio}},z_{\text{vio}},\psi_{\text{vio}})^\intercal - (x_{\text{ld}},y_{\text{ld}},z_{\text{ld}},\psi_{\text{ld}})^\intercal.
\end{align}
The onboard camera measurement system typically operates at a slower rate than \ac{VIO}.
In our case, \ac{VIO} runs at \SI{100}{\hertz} and the Landmark detector runs at \SI{30}{\hertz}.
To align these states in time, we store the last \SI{1}{\second} of incoming \ac{VIO} odometry in a buffer and interpolate between the samples whose timestamps are closest to the camera measurement.

The measurement equation defines how the system states are observed through available measurements.
The model $\vec{z}_d$ for our measurements $\vec{z}$ at time $t$ is given by
\begin{align}
    \vec{z}_d(t) &= \vec{H}_d\vec{x}_d(t) + \vec{v}_{\text{noise}}(t),\\[0.2cm]
    \vec{H}_d &= \begin{pmatrix}
        1 & 0 & 0 & 0 & 0 & 0 & 0 & 0\\
        0 & 0 & 1 & 0 & 0 & 0 & 0 & 0\\
        0 & 0 & 0 & 0 & 1 & 0 & 0 & 0\\
        0 & 0 & 0 & 0 & 0 & 0 & 1 & 0
    \end{pmatrix},
\end{align}
where the vector $\vec{v}_{\text{noise}}$ represents measurement noise introduced by sensing uncertainties and errors.
Similar to $\vec{w}_{\text{noise}}$, each component of $\vec{v}_{\text{noise}}$ is modeled as independent Gaussian noise with zero mean and covariance values $R_{x_d}$, $R_{y_d}$, $R_{z_d}$, $R_{\psi_d}$ $\in \mathbb{R}^+$.

We use a \ac{LKF} as a state estimator for \ac{VIO} drift.
The \ac{LKF} consists of two steps: 1) the prediction step and 2) the correction step.
In the prediction step, the system model is used to propagate the current state estimate and the corresponding covariance matrix $\vec{P}_d$ of the state $\vec{x}_d$ forward in time
\begin{align}
    \vec{x}_d(t+\dt{}) &= \vec{A}_d\vec{x}_d(t),\\
    \vec{P}_d(t+\dt{}) &= \vec{A}_d\vec{P}_d(t)\vec{A}_d^\intercal + \vec{Q}_d,
\end{align}
where the covariance matrix $\vec{Q}_d$ is a diagonal matrix
\begin{align}
    \vec{Q}_d = \text{diag}\{ Q_{x_d}, Q_{y_d}, Q_{z_d}, Q_{\psi_d}, Q_{\dot{x}_d}, Q_{\dot{y}_d}, Q_{\dot{z}_d}, Q_{r_d} \}.
\end{align}
The prediction step runs at \SI{100}{\hertz}, resulting in $\dt{}=\SI{0.01}{\second}$.
In the correction step, the current estimated state is updated using the incoming measurement $\vec{z}(t)$
\begin{align}
    \vec{G}_d(t) &\mathrel{:=} \vec{P}_d(t)\vec{H}_d(t)^\intercal\left( \vec{H}_d \vec{P}_d(t) \vec{H}_d^\intercal + \vec{R}_d \right)^{-1},\\
    \vec{x}_d(t) &\mathrel{:=} \vec{x}_d(t) + \vec{G}_d(t)\left( \vec{z}(t) - \vec{H}_d\vec{x}_d(t) \right),\\
    \vec{P}_d(t) &\mathrel{:=} \vec{P}_d(t) - \vec{G}_d(t)\vec{H}_d\vec{P}_d(t),
\end{align}
where $\vec{R}_d$ is defined as a diagonal matrix
\begin{align}
    \vec{R}_d = \text{diag} \{ R_{x_d}, R_{y_d}, R_{z_d}, R_{\psi_d} \}.
\end{align}

The covariances $R_{x_d}$, $R_{y_d}$, $R_{z_d}$, and $R_{\psi_d}$ are estimated using the confidence $c_{\text{ld}}$ of the Landmark detector \refeq{eq:confidence} as follows
\begin{align}
     R_{\{x_d,y_d,z_d,\psi_d\}} = \frac{a_{rv}}{1 + \exp(-b_{rv}(c_{\text{ld}} - c_{rv}))} + d_{rv}, \label{eq:covariance}
\end{align}
where $a_{rv}$, $b_{rv}$, $c_{rv}$, and $d_{rv}$ are parameters in $\mathbb{R}^+$.
Unlike \cite{kaufmann2023ChampionlevelDroneRacing}, which estimates covariance through multi-run PnP executions, the closed-form equation \refeq{eq:covariance} evaluates in constant time.
Multi-run PnP is computationally demanding and significantly increases processing times, which is unfeasible on resource-constrained platforms during high-speed flight, where state estimation delays lead to degraded control performance.
Since the \ac{LKF} is sensitive to outliers, we reject measurements deemed unreliable based on the confidence $c_{\text{ld}}$ defined in \refeq{eq:confidence}. 
If the measurement confidence $c_{\text{ld}}$ falls below a predefined threshold $t_{c_{\text{ld}}}=0.5$, the measurement is not fused.

\subsection{State estimator}
\label{sec:state_estimator}
All data from the \ac{VIO} algorithm, the estimated \ac{VIO} drift, and the \ac{IMU} measurements are fused in the State estimator (see \reffig{fig:diagram}).
The corrected \ac{UAV} position $\vec{p}_{\text{uav}}$ is computed as the difference between the \ac{VIO} position transformed into the $\worldframe{}$ frame $\vec{p}_{\text{vio}}^{w}$ and the estimated position drift $\vec{p}_d$ as
\begin{align}
    \vec{p}_{\text{uav}}(t) =  \vec{p}_{\text{vio}}^{w}(t) - \vec{p}_d(t).
\end{align}
The roll $\phi_{\text{vio}}$ and pitch $\theta_{\text{vio}}$ obtained from the \ac{VIO} algorithm are inaccurate and delayed due to z-axis alignment issues and the lack of incorporation into the graph optimization \cite{qin2018VINSMonoRobustVersatile}.
Therefore, the roll angle $\phi_{\text{imu}}$ and pitch angle $\theta_{\text{imu}}$ are taken from the \ac{IMU} data, which are filtered using a low-pass filter and a dynamic notch filter to ensure fast and accurate response during aggressive maneuvers in agile flights.
The yaw angle is taken from the \ac{VIO} algorithm, as it is optimized through global pose graph optimization, which makes it more accurate compared to the yaw $\psi_{\text{imu}}$ derived from the \ac{IMU}.
To further reduce yaw drift, the estimated yaw drift $\psi_d$ is used to correct the \ac{VIO} yaw angle $\psi_{\text{vio}}$ in $\worldframe{}$ frame.
The correction is applied in quaternion form as
\begin{align}
    \mathbf{q}_{\psi_{\text{uav}}}(t) &= \mathbf{q}_{\text{d}}^{-1}(t) \otimes \mathbf{q}_{\text{vio}}(t),
\end{align}
where $ \mathbf{q}_{\text{d}}$ denotes the estimated yaw drift, $\mathbf{q}_{\text{vio}}$ represents \ac{VIO} orientation, and $\mathbf{q}_{\psi_{\text{uav}}}$ is drift-compensated \ac{VIO} orientation.
The corrected yaw angle, $\psi_{\text{uav}}$, is subsequently extracted from $\mathbf{q}_{\psi_{\text{uav}}}$.

The linear velocities are taken from the \ac{VIO} algorithm and corrected using the estimated drift velocities.
We assume that the velocity error of the VIO estimator is approximately consistent with the time derivative of the position error.
The linear velocity from the \ac{VIO} algorithm is expressed in the $\viobodyframe{}$ frame, whereas the drift linear velocity is estimated in the $\worldframe{}$ frame.
Therefore, the \ac{VIO} velocities are transformed from the $\viobodyframe{}$ frame to the $\worldframe{}$ frame using the rotation matrix $\vec{R}_{\viobodyframe{}}^{\worldframe{}}$
\begin{align}
    \vec{v}_{\text{vio}}^{w}(t) &= \vec{R}_{\viobodyframe{}}^{\worldframe{}} \vec{v}_{\text{vio}}^{\viobodyframe{}}(t),\\
    \vec{v}_{\text{uav}}(t) &= \vec{v}_{\text{vio}}^{w}(t) - \vec{v}_d(t).
\end{align}
The roll rate $p_{\text{imu}}$ and pitch rate $q_{\text{imu}}$ are taken directly from the \ac{IMU}.
However, the yaw rate $r_{\text{vio}}$ from the \ac{VIO} algorithm is corrected using the estimated drift rate $r_d$
\begin{align}
    r_{\text{uav}} = r_{\text{vio}} - r_d.
\end{align}

To summarize, the corrected \ac{UAV} position $\vec{p}_{\text{uav}}$, orientation $\vec{\Theta}_{\text{uav}}$, linear velocity $\vec{v}_{\text{uav}}$, and angular velocity $\bm{\omega}_{\text{uav}}$ are computed as follows
\begin{align}
    \vec{p}_{\text{uav}}(t) &=  \vec{p}_{\text{vio}}^w(t) - \vec{p}_d(t),\\
    \vec{\Theta}_{\text{uav}}(t) &= (\phi_{\text{imu}}(t),\theta_{\text{imu}}(t), \psi_{\text{uav}})^\intercal,\\
    \vec{v}_{\text{uav}}(t) &= \vec{v}_{\text{vio}}^{w}(t) - \vec{v}_d(t),\\
    \bm{\omega}_{\text{uav}}(t) &= (p_{\text{imu}}(t), q_{\text{imu}}(t), r_{\text{vio}}(t) - r_d(t))^\intercal.   
\end{align}
This estimation pipeline enables fast flights through gates in drone racing, providing accurate \ac{UAV} states even during aggressive maneuvers.
Our method uses \ac{VIO} odometry, which does not require predefined landmarks and is subsequently corrected using the estimated drift states.
During fast and aggressive flights, the drift changes dynamically, and our estimation approach captures these variations through the drift velocity estimation.
Due to the drift velocity estimation, our approach maintains accurate state estimation even when the Landmark detector temporarily cannot provide measurements, for example, when landmarks are not visible in the camera frame, or they are too distant from the camera, which negatively influences the precision of the onboard camera measurement system.
The direct fusion of \ac{IMU} data enables fast and accurate estimation of roll, pitch, and their corresponding rates during aggressive maneuvers.

\section{RESULTS}
\label{sec:results}
The proposed method was first evaluated in Gazebo simulations and subsequently tested in real-world experiments and a racing competition\textsuperscript{\ref{a2rl_challenge}}.
The gates are square-shaped, with an inner opening of \SI{1.5}{\meter} and an outer dimension of \SI{2.7}{\meter}.
The UAV, including its propellers, fits within a bounding box of dimensions $0.4$$\times$$0.4$$\times$\SI{0.25}{\meter}.

\subsection{Gazebo simulations}
We ran 1600 simulations to evaluate the robustness of our approach with respect to delay $\tau_{d}$ and noise in the Landmark detector.
The noise was modeled as Gaussian with zero mean and standard deviation equal to $\sigma^2_{\vec{p}_{\text{ld}}}$ for the position noise and $\sigma^2_{\psi_{\text{ld}}}$ for the yaw noise.
For each combination of $\tau_{d} = \{0, 15, 30, 60\}~\SI{}{\milli\second}$ with $\sigma^2_{\vec{p}_{\text{ld}}} = \{0.001, 0.01, 0.1, 1.0\}~\SI{}{\meter}$ and $\sigma^2_{\psi_{\text{ld}}} = \{0.001, 0.01, 0.1, 1.0\}~\SI{}{\radian}$, we conducted 100 simulations and computed the mean \ac{RMSE} for individual \ac{UAV} states for both our approach and \ac{VIO}, relative to the ground-truth values.
The results are summarized in \reftab{tab:rmse_sim} and show that our system operates effectively even in the presence of delays and noise in the Landmark detector.
A significant increase in \ac{RMSE} occurs for all states when $\sigma^2_{\vec{p}_{\text{ld}}}$ and $\sigma^2_{\psi_{\text{ld}}}$ reach 1.0.
At this level, the linear velocity estimates from our approach and \ac{VIO} exhibit similar precision.
In all other scenarios, our approach outperforms the baseline \ac{VIO} across all state estimates (see \reftab{tab:rmse_sim}).
Specifically, because \cite{foehn2022AlphaPilotAutonomousDrone} leaves linear velocity $\vec{v}$ and angular velocity $\bm{\omega}$ uncorrected, it directly inherits the higher \ac{VIO} velocity errors reported in \reftab{tab:rmse_sim}.
In contrast, our approach corrects these states and thus achieves better velocity accuracy.
Similarly, because \cite{kaufmann2023ChampionlevelDroneRacing} directly uses raw \ac{VIO} attitude $\vec{\Theta}$ and angular rates $\bm{\omega}$, it inherits the baseline \ac{VIO} orientation and angular rate errors from \reftab{tab:rmse_sim}, whereas our state estimation pipeline yields significantly lower attitude and angular rate errors.

\begin{table}[t]
\caption{RMSE results from the simulations.}
\setlength{\tabcolsep}{4pt}
\scriptsize
\centering
\label{tab:rmse_sim}
\begin{tabular}{ll|cccc|cccc}
\hline
\multicolumn{2}{l|}{$\bm{\sigma^2_{\vec{p}_{\text{ld}}}}$ [\SI{}{\meter}]} & \bf 0.001 & \bf 0.01 & \bf 0.1 & \bf 1.0 & \bf 0.001 & \bf 0.01 & \bf 0.1 & \bf 1.0\\[0.1cm]
\multicolumn{2}{l|}{$\bm{\sigma^2_{\psi_{\text{ld}}}}$ [\SI{}{\radian}]} & \bf 0.001 & \bf 0.01 & \bf 0.1 & \bf 1.0 & \bf 0.001 & \bf 0.01 & \bf 0.1 & \bf 1.0\\\hline
\makecell{$\bm{\tau_{\text{\textbf{d}}}}$\\\text{[\SI{}{\milli\second}]}} & \multicolumn{1}{c}{\textbf{State}} & \multicolumn{4}{c}{\textbf{Our approach}} & \multicolumn{4}{c}{\textbf{VIO}}\\\hline
\multirow{4}{*}{\bf{0}} & $\vec{p}$ [\SI{}{\meter}] & \bf 0.35 & 0.40 & 0.45 & 1.11 & 6.39 & 7.55 & 8.68 & 8.44 \\
 & $\vec{\Theta}$ [\SI{}{\radian}] & \bf 0.04 & 0.04 & 0.09 & 0.57 & 0.23 & 0.23 & 0.26 & 0.29 \\
 & $\vec{v}$ [\SI{}{\meter\per\second}] & \bf 0.96 & 1.00 & 1.09 & 1.40 & 1.27 & 1.30 & 1.49 & 1.57 \\
 & $\bm{\omega}$ [\SI{}{\radian\per\second}] & 0.24 & \bf 0.23 & 0.24 & 0.24 & 1.13 & 1.15 & 1.11 & 1.09 \\
\hline
\multirow{4}{*}{\bf{15}} & $\vec{p}$ [\SI{}{\meter}] & \bf 0.23 & 0.47 & 0.32 & 1.19 & 3.94 & 8.77 & 5.63 & 9.39 \\
 & $\vec{\Theta}$ [\SI{}{\radian}] & \bf 0.04 & 0.04 & 0.09 & 0.57 & 0.09 & 0.23 & 0.14 & 0.32 \\
 & $\vec{v}$ [\SI{}{\meter\per\second}] & \bf 0.69 & 1.08 & 0.81 & 1.49 & 0.76 & 1.52 & 0.98 & 1.68 \\
 & $\bm{\omega}$ [\SI{}{\radian\per\second}] & \bf 0.23 & 0.23 & 0.24 & 0.23 & 1.21 & 1.14 & 1.18 & 1.09 \\
\hline
\multirow{4}{*}{\bf{30}} & $\vec{p}$ [\SI{}{\meter}] & 0.55 & \bf 0.48 & 0.59 & 0.88 & 6.96 & 7.62 & 7.64 & 4.17 \\
 & $\vec{\Theta}$ [\SI{}{\radian}] & \bf 0.04 & 0.04 & 0.10 & 0.56 & 0.25 & 0.18 & 0.28 & 0.10 \\
 & $\vec{v}$ [\SI{}{\meter\per\second}] & 1.01 & \bf 0.92 & 0.94 & 1.08 & 1.34 & 1.26 & 1.46 & 0.84 \\
 & $\bm{\omega}$ [\SI{}{\radian\per\second}] & 0.24 & \bf 0.23 & 0.24 & 0.24 & 1.14 & 1.15 & 1.10 & 1.20 \\
\hline
\multirow{4}{*}{\bf{60}} & $\vec{p}$ [\SI{}{\meter}] & 0.72 & \bf 0.66 & 0.73 & 1.09 & 7.36 & 7.11 & 7.72 & 6.51 \\
 & $\vec{\Theta}$ [\SI{}{\radian}] & \bf 0.05 & 0.05 & 0.10 & 0.56 & 0.25 & 0.21 & 0.23 & 0.18 \\
 & $\vec{v}$ [\SI{}{\meter\per\second}] & 1.06 & \bf 0.97 & 1.06 & 1.13 & 1.40 & 1.26 & 1.40 & 1.19 \\
 & $\bm{\omega}$ [\SI{}{\radian\per\second}] & 0.24 & \bf 0.23 & 0.23 & 0.24 & 1.14 & 1.15 & 1.14 & 1.16 \\
\hline
\end{tabular}
\end{table}

\subsection{Real-world experiments}
\begin{figure}[!b]
  \centering
  \includegraphics[width=\linewidth]{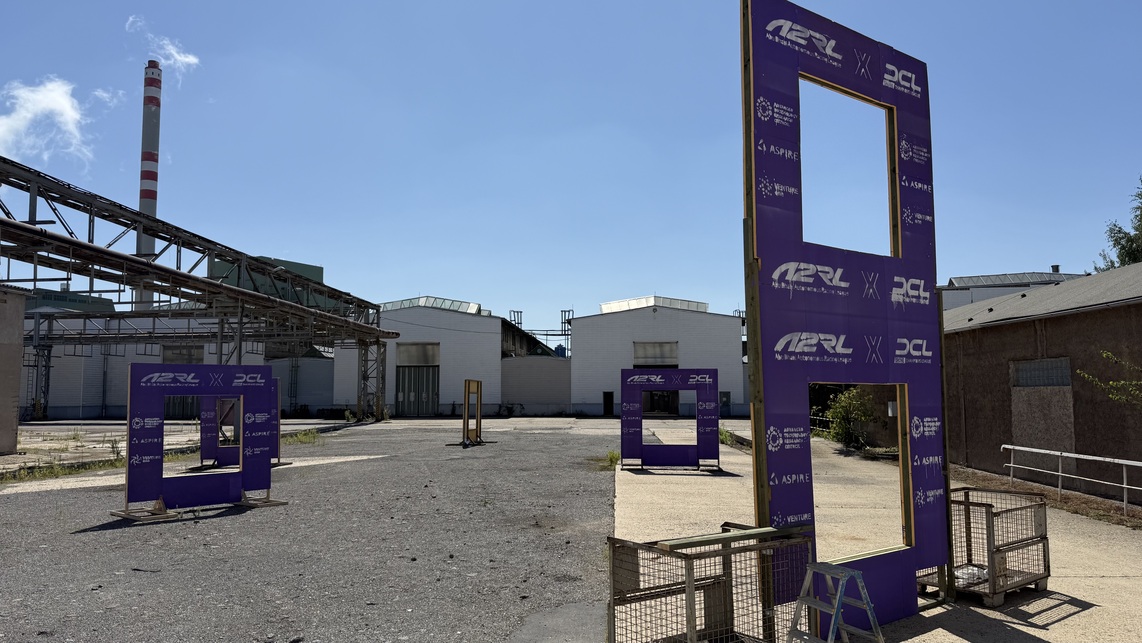}
  \caption{Custom-built outdoor track used to verify our approach.}
  \label{fig:kladno_race_track}
\end{figure}

To evaluate the proposed method against ground-truth data, we conducted real-world experiments on an outdoor race track (see \reffig{fig:kladno_race_track}) consisting of 5 gates, as illustrated in \reffig{fig:2d_plot_kladno}.
The fourth gate was designed as a double gate, where the \ac{UAV} performed a Split-S maneuver from the top to the bottom gate.
Ground-truth data were obtained using a \ac{GPS}-\ac{RTK} system.
The \ac{UAV} was equipped with a Holybro Pixhawk 6C Mini flight controller and a Holybro H-RTK F9P Helical \ac{GPS}-\ac{RTK} module.

\addtocounter{figure}{1}
\begin{figure*}[!tb]
  \centering
  \def\svgwidth{0.8\linewidth}
  \include{figure6.tex}
  \vspace{-1.0cm}
  \caption{3D plot of the estimated UAV position using our approach, with color indicating the UAV speed. The VIO position is shown in grey. Measurements from the Landmark detector are represented as circles, with green color indicating confidence, and the race gates are shown as black circles.}
  \label{fig:3d_plot_uae}
\end{figure*}

\addtocounter{figure}{-2}
\begin{figure}[!t]
  \centering
  \def\svgwidth{\linewidth}
  \include{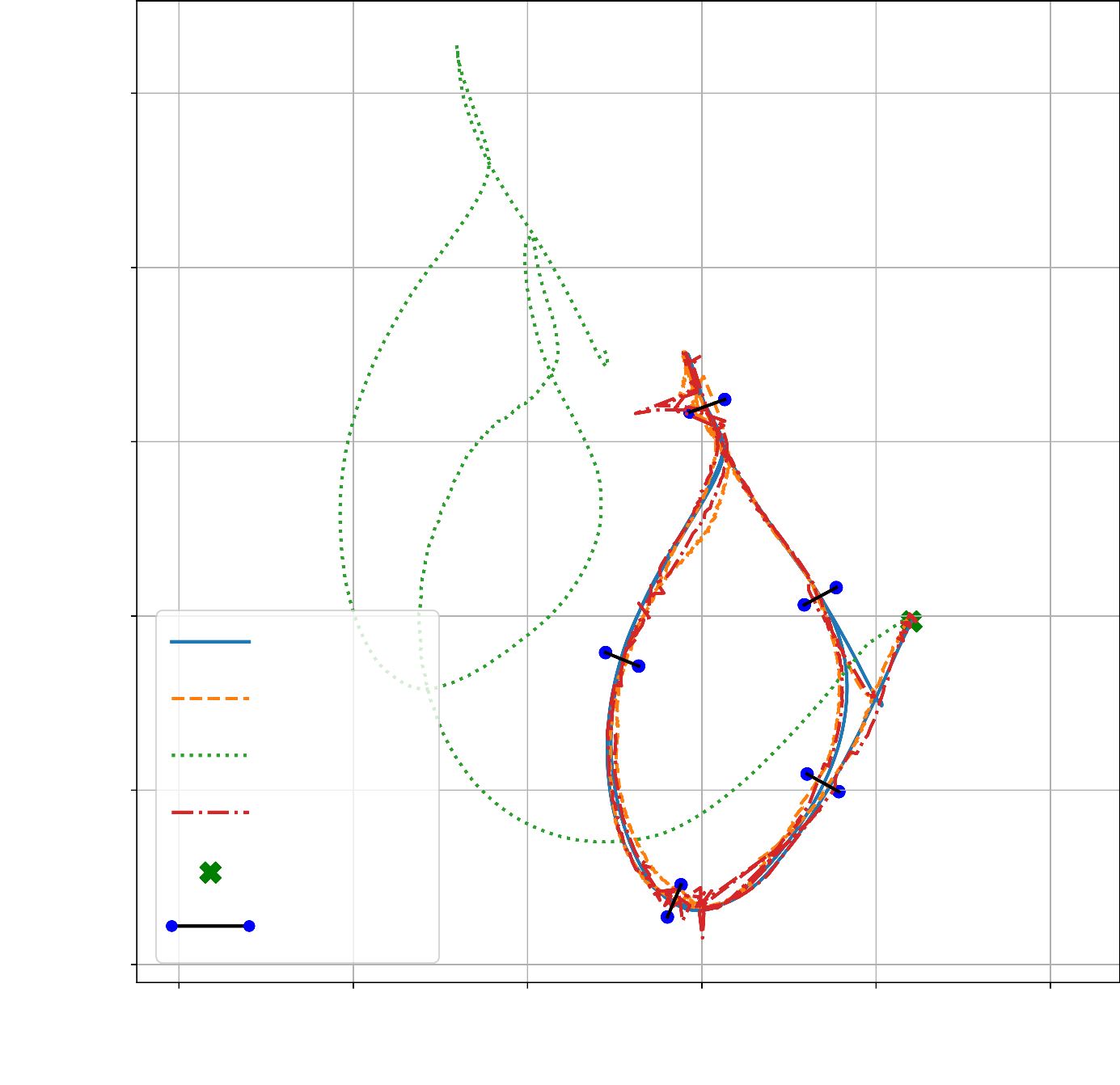}
  \vspace{-1.0cm}
  \caption{Top view of the outdoor track showing the \ac{RTK} states, results of our approach, \ac{VIO} states, and data from the Landmark detector (LM det.).}
  \label{fig:2d_plot_kladno}
\end{figure}
\addtocounter{figure}{1}

In each of our flight tests, we performed 2 laps on the track, running both our proposed estimation approach and the \ac{RTK}-based state estimator simultaneously for comparison.
During these experiments, the \ac{UAV} reached speeds of up to \SI{10}{\meter\per\second}.
\reffigfull{fig:2d_plot_kladno} presents a 2D plot of the track in the $x$–$y$ plane, showing the gate positions, podium position, \ac{RTK} states, results of our approach (\refsec{sec:state_estimator}), \ac{VIO} states (\refsec{sec:vio_algorithm}), and outputs from the Landmark detector (\refsec{sec:gate_estimator}).
As illustrated, the \ac{VIO} estimates (shown in green) began drifting shortly after the start of the trajectory.
The measurements from the Landmark detector are highly noisy and exhibit significant spikes.
In contrast, the states estimated by our approach follow the \ac{RTK} ground-truth data throughout the flight.

\reftabfull{tab:kladno_overall_error_stats} shows the \ac{RMSE}, mean error, maximum error, and standard deviation of the error for position $\vec{p}$, orientation $\vec{\Theta}$, linear velocity $\vec{v}$, and angular velocity $\bm{\omega}$.
Since the Landmark detector provides only position and orientation measurements, \ac{RMSE} for linear and angular velocities is not reported.
The results confirm that \ac{VIO} alone is insufficient to complete the race track due to accumulated drift.
Indeed, standalone \ac{VIO} yields the worst performance metrics across all estimated states, whereas our approach significantly improves estimation accuracy (\reftab{tab:kladno_overall_error_stats}).

\begin{table}[!htb]
\centering
\caption{RMSE, mean error, standard deviation of the error, and maximum error of estimated UAV states using our approach compared to state-of-the-art, VIO, and Landmark detector.}
\label{tab:kladno_overall_error_stats}
\scriptsize
\begin{tabular}{llcccc}
\toprule
\textbf{Method} & \textbf{State} & \textbf{RMSE} & \textbf{Mean} & \textbf{Std. Dev.} & \textbf{Max} \\
\hline
\multirow{4}{*}{\shortstack[l]{\bf Our\\\bf approach}}
 & $\vec{p}$ [\SI{}{\meter}]         & 0.543 & 0.477 & 0.260 & 1.490 \\
 & $\vec{\Theta}$ [\SI{}{\radian}]     & 0.092 & 0.077 & 0.050 & 0.229 \\
 & $\vec{v}$ [\SI{}{\meter\per\second}]  & 1.262 & 1.036 & 0.722 & 3.689 \\
 & $\bm{\omega}$ [\SI{}{\radian\per\second}] & 0.103 & 0.074 & 0.071 & 0.709 \\
\hline
\multirow{4}{*}{\shortstack[l]{Foehn et al. \cite{foehn2022AlphaPilotAutonomousDrone}}}
 & $\vec{p}$ [\SI{}{\meter}]         & 0.645 & 0.535 & 0.359 & 2.396 \\
  & $\vec{\Theta}$ [\SI{}{\radian}]      & 0.179 & 0.163 & 0.074 & 0.343 \\
  & $\vec{v}$ [\SI{}{\meter\per\second}]  & 1.539 & 1.230 & 0.924 & 4.536 \\
 & $\bm{\omega}$ [\SI{}{\radian\per\second}] & 0.924 & 0.688 & 0.616 & 3.470 \\
\hline
\multirow{4}{*}{\shortstack[l]{Kaufmann et al. \cite{kaufmann2023ChampionlevelDroneRacing}}}
 & $\vec{p}$ [\SI{}{\meter}]         & 0.554 & 0.495 & 0.249 & 1.464 \\
  & $\vec{\Theta}$ [\SI{}{\radian}]  & 0.212 & 0.204 & 0.057 & 0.342  \\
  & $\vec{v}$ [\SI{}{\meter\per\second}]  & 1.327 & 1.127 & 0.701 & 4.314 \\
 & $\bm{\omega}$ [\SI{}{\radian\per\second}] & 0.924 & 0.688 & 0.616 & 3.470 \\
\hline
\multirow{4}{*}{\shortstack[l]{VIO}}
 & $\vec{p}$ [\SI{}{\meter}]         & 16.734 & 14.990 & 7.438 & 26.908 \\
  & $\vec{\Theta}$ [\SI{}{\radian}]      & 0.212 & 0.204 & 0.057 & 0.342 \\
  & $\vec{v}$ [\SI{}{\meter\per\second}]  & 1.539 & 1.230 & 0.924 & 4.536 \\
 & $\bm{\omega}$ [\SI{}{\radian\per\second}] & 0.924 & 0.688 & 0.616 & 3.470 \\
\hline
\multirow{2}{*}{\shortstack[l]{Landmark\\detector}}
 & $\vec{p}$ [\SI{}{\meter}]         & 0.608 & 0.461 & 0.396 & 7.192 \\
 & $\vec{\Theta}$ [\SI{}{\radian}]      & 0.182 & 0.124 & 0.133 & 1.074 \\
\bottomrule
\end{tabular}
\end{table}

Because the approach in \cite{kaufmann2023ChampionlevelDroneRacing} directly uses orientation $\vec{\Theta}$ and angular velocity $\bm{\omega}$ from the \ac{VIO}, our framework significantly reduces the \ac{RMSE} of $\vec{\Theta}$ by \SI{57}{\percent} and the \ac{RMSE} of $\bm{\omega}$ by a factor of 9.
Furthermore, our method yields lower \ac{RMSE} values for both position and linear velocity compared to \cite{kaufmann2023ChampionlevelDroneRacing}.
When compared to \cite{foehn2022AlphaPilotAutonomousDrone}, which directly utilizes $\vec{v}$ and $\bm{\omega}$ from the \ac{VIO}, our approach reduces the linear velocity \ac{RMSE} by \SI{18}{\percent} and the angular velocity \ac{RMSE} by a factor of 9, while also achieving smaller errors in position and orientation.
The mean error across all states further support this comparison with state-of-the-art methods (see \reftab{tab:kladno_overall_error_stats}).

\subsection{A2RL Drone Racing Challenge}
\label{sec:results_racing_challenge}
The \ac{UAV} used during A2RL Drone Racing Challenge\textsuperscript{\ref{a2rl_challenge}} was equipped with an 8 MP camera, an onboard computer, and a flight controller with an \ac{IMU}.
The track consisted of 10 gates, with 1 double gate in which split-S maneuver was required.
The teams had to fly 2 laps through gates.
The team with the shortest time won, and the time was measured from passing the first gate to passing the last gate in the second lap.
The part of the track is shown in \reffig{fig:flight_long_exposure}, and the \ac{UAV} flying in it is depicted in \reffig{fig:diagram}.
In each round, the team had a 15-minute time window to perform as many flights as desired, with the fastest time recorded for the leaderboard of that round.

\begin{figure}[!tb]
  \centering
  \def\svgwidth{\linewidth}
  \include{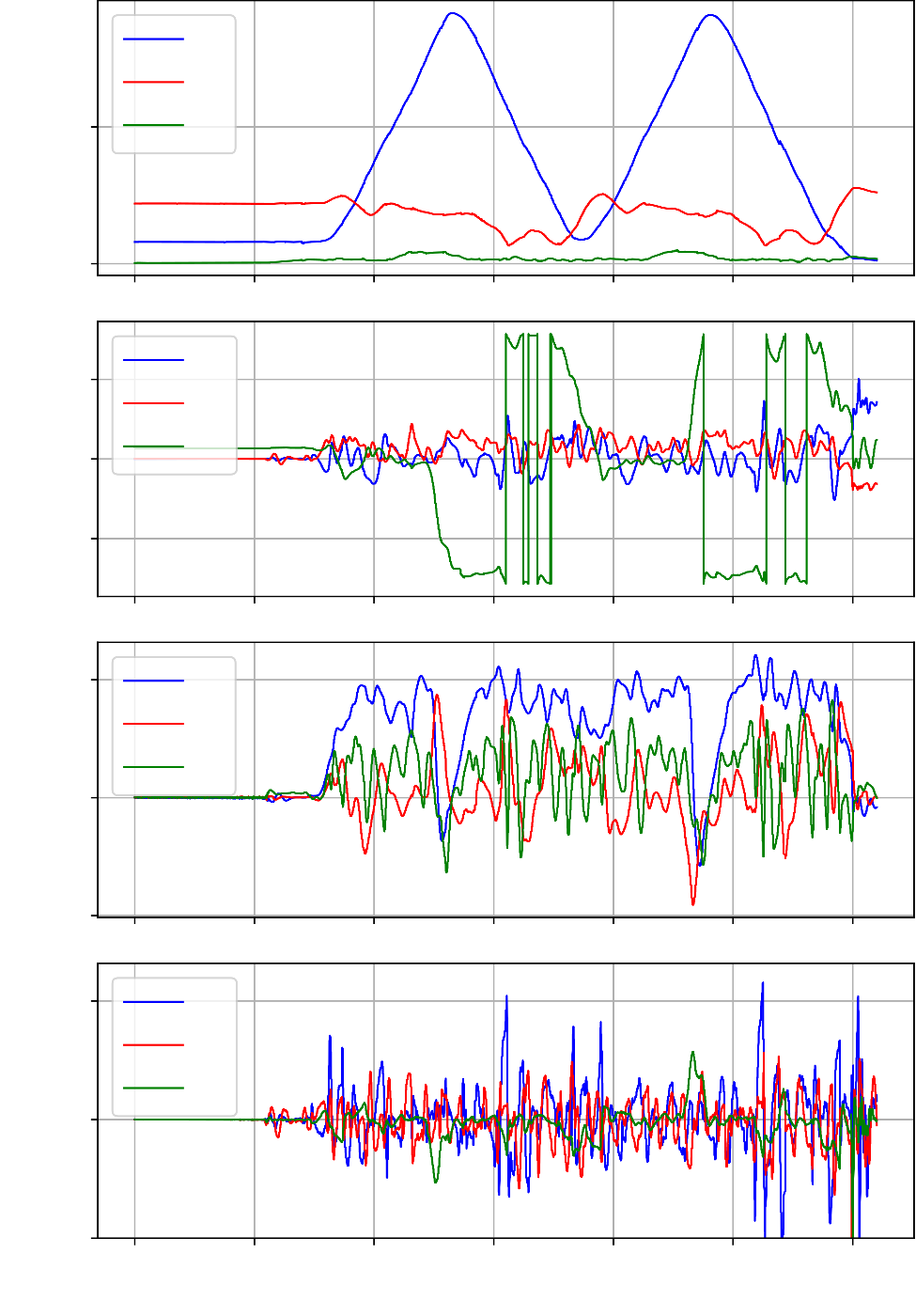}
  \vspace{-0.9cm}
  \caption{Estimated UAV states during the A2RL finals using our approach.}
  \label{fig:uav_states}
\end{figure}

We completed 2 laps multiple times at speeds ranging from \SI{5}{\meter\per\second} to \SI{12.5}{\meter\per\second} (see video\footnote{\href{https://mrs.fel.cvut.cz/papers/onboard-uav-state-estimation-a2rl}{https://mrs.fel.cvut.cz/papers/onboard-uav-state-estimation-a2rl}}).
An example of our fastest trajectory is shown in \reffig{fig:3d_plot_uae}, which clearly demonstrates that the \ac{VIO} algorithm alone is insufficient for drone racing.
\reffigfull{fig:uav_states} shows the estimated individual \ac{UAV} states $\vec{x}_{\text{uav}}$ obtained using our approach.
Using the presented estimation approach, our team successfully advanced through the qualification, quarterfinal, and semifinal rounds, reaching the finals alongside three other teams out of a total of 210 and winning a medal.
These results from such a challenging competition further support that our estimation approach is able to maintain great performance even during fast and aggressive flight maneuvers.

\section{CONCLUSION}
In this paper, we presented an approach for \ac{GNSS}-denied \ac{UAV} state estimation during fast vision-based flights and aggressive maneuvers in cluttered environments.
Our method estimates \ac{UAV} states in 6 \acp{DOF} (position, orientation, linear velocity, and angular velocity) using only a single RGB monocular camera, an \ac{IMU}, and an onboard computer.
It builds upon a \ac{VIO} method that provides drifting and inaccurate states, which are corrected using measurements from the onboard landmark-based camera system.
Our approach fuses data from \ac{VIO}, the onboard landmark-based camera measurement system, and the \ac{IMU} to estimate and compensate for \ac{VIO} drift, correcting the \ac{UAV}’s position, orientation, and linear and angular velocities.
For sensor fusion, we introduced a novel mathematical model of the drift.
Additionally, directly incorporating \ac{IMU} data into the final estimate ensures fast response and high precision even during highly aggressive maneuvers.
Unlike end-to-end solutions, our method is well-suited for industrial and safety-critical applications, as it decomposes the problem into planning, state estimation, and control, where theoretical guarantees exist.
The approach was extensively evaluated in simulations and real-world experiments, outperforming current state-of-the-art methods.
Finally, we applied the proposed approach during the A2RL Drone Racing Challenge 2025\textsuperscript{\ref{a2rl_challenge}}, where our team reached the final four out of 210 teams.

\bibliographystyle{elsarticle-num}
\bibliography{main.bib}

\begin{acronym}
    \acro{VIO}{Visual-Inertial Odometry}
    \acro{VO}{Visual Odometry}
    \acro{ESC}{Electronic speed controller}
    \acro{VINS}{Visual-Inertial Navigation System}
  \acro{CNN}[CNN]{Convolutional Neural Network}
  \acro{IR}[IR]{infrared}
  \acro{GNSS}[GNSS]{Global Navigation Satellite System}
  \acro{MOCAP}[mo-cap]{Motion capture}
  \acro{MPC}[MPC]{Model Predictive Control}
  \acro{MRS}[MRS]{Multi-robot Systems group}
  \acro{ML}[ML]{Machine Learning}
  \acro{MAV}[MAV]{Micro-scale Unmanned Aerial Vehicle}
  \acro{UAV}[UAV]{Unmanned Aerial Vehicle}
  \acro{UV}[UV]{ultraviolet}
  \acro{UVDAR}[\emph{UVDAR}]{UltraViolet Direction And Ranging}
  \acro{UT}[UT]{Unscented Transform}
  \acro{RTK}[RTK]{Real-Time Kinematic}
  \acro{ROS}[ROS]{Robot Operating System}
  \acro{wrt}[w.r.t.]{with respect to}
  \acro{LTI}[LTI]{Linear time-invariant}
  \acro{USV}[USV]{Unmanned Surface Vehicle}
  \acroplural{DOF}[DOFs]{Degrees of Freedom}
  \acro{DOF}[DOF]{Degree of Freedom}
  \acro{API}[API]{Application Programming Interface}
  \acro{CTU}[CTU]{Czech Technical University}
  \acroplural{DOF}[DOFs]{Degrees of Freedom}
  \acro{DOF}[DOF]{Degree of Freedom}
  \acro{FOV}[FOV]{Field of View}
  \acro{GNSS}[GNSS]{Global Navigation Satellite System}
  \acro{GPS}[GPS]{Global Positioning System}
  \acro{IMU}[IMU]{Inertial Measurement Unit}
  \acro{LKF}[KF]{Kalman Filter}
  \acro{KF}[KF]{Kalman Filter}
  \acro{LTI}[LTI]{Linear time-invariant}
  \acro{LiDAR}[LiDAR]{Light Detection and Ranging}
  \acro{MAV}[MAV]{Micro Aerial Vehicle}
  \acro{MPC}[MPC]{Model Predictive Control}
  \acro{MRS}[MRS]{Multi-robot Systems}
  \acro{ROS}[ROS]{Robot Operating System}
  \acro{RTK}[RTK]{Real-time Kinematics}
  \acro{SLAM}[SLAM]{Simultaneous Localization And Mapping}
  \acro{UAV}[UAV]{Unmanned Aerial Vehicle}
  \acro{UGV}[UGV]{Unmanned Ground Vehicle}
  \acro{UKF}[UKF]{Unscented Kalman Filter}
  \acro{USV}[USV]{Unmanned Surface Vehicle}
  \acro{RMSE}[RMSE]{Root Mean Square Error}
  \acro{UVDAR}[UVDAR]{UltraViolet Direction And Ranging}
  \acro{UV}[UV]{UltraViolet}
  \acro{VRX}[VRX]{Virtual RobotX}
  \acro{WAM-V}[WAM-V]{Wave Adaptive Modular Vessel}
  \acro{GT}[GT]{Ground Truth}
  \acro{UTM}[UTM]{Universal Transverse Mercator}
\end{acronym}

\end{document}